\documentclass[letterpaper, 10 pt, conference]{ieeeconf}
\usepackage[T1]{fontenc}    
\usepackage[utf8]{inputenc} 

\usepackage{url}            
\usepackage{booktabs}       
\usepackage{amsfonts}       

\usepackage[dvipsnames,svgnames]{xcolor}

\usepackage{amsmath} 

\usepackage{amsthm}
\usepackage{amssymb} 
\usepackage{mathtools}
\usepackage{mathrsfs}
\usepackage{mathdots}
\usepackage{bm}
\usepackage[linesnumbered,ruled,vlined]{algorithm2e}

\usepackage{graphics} 
\usepackage{tikz}
\usetikzlibrary{shapes, shapes.geometric, arrows, calc}
\usetikzlibrary{bayesnet}
\usetikzlibrary{fadings}
\usetikzlibrary{patterns}
\usepackage{yhmath}
\usepackage{cancel}
\usepackage{gensymb}

\usepackage{pgfkeys}
\usepackage{standalone}
\usepackage{fontawesome5} 

\usepackage{subcaption}
\usepackage{cite}
\usepackage{graphicx}
\usepackage{multirow}
\usepackage{threeparttable}
\usepackage{adjustbox}
\usepackage{aligned-overset}
\usepackage{dblfloatfix}

\usepackage{float}
\usepackage{verbatim}
\let\labelindent\relax
\usepackage[shortlabels]{enumitem}
\usepackage{array}

\usepackage{hyperref} 
\hypersetup{
 colorlinks = true,
 linkcolor=black,
 urlcolor=Blue}

\definecolor{revise}{rgb}{0.54, 0.17, 0.89} 

\IEEEoverridecommandlockouts

\title{Robotic Boomerang Throwing via Model-Based Release Design}
\author{Yang Liu$^{1}$ \quad Colin Jones$^{2}$ \quad Aude Billard$^{1}$%
\thanks{$^{1}$Yang Liu and Aude Billard are with the Learning Algorithms and Systems Lab (LASA), EPFL, Switzerland.}%
\thanks{$^{2}$Colin Jones is with the Automatic Control Lab, EPFL, Switzerland.}%
\thanks{Corresponding author: Yang Liu ({\tt\footnotesize yangliudh@gmail.com}).}%
}

\begin{document}
\maketitle

\begin{abstract}
Throwing objects that generate aerodynamic lift can greatly extend robot throwing beyond ballistic flight. A returning boomerang is a challenging example because its flight depends strongly on the release velocity, attitude, and spin, while robotic manipulators cannot readily reproduce the rapid motions used in human throwing. We present a model-based framework for robotic boomerang throwing centered on the release state. We identify the boomerang flight dynamics in stages to predict how flight changes across design variations. To systematically design the robot throwing motion, we screen candidate parameters according to how strongly and consistently they control release spin under uncertain contact conditions. These models are then used to design the throwing motion and boomerang for a 6-DoF manipulator with limited joint speeds. To our knowledge, this is the first robotic manipulator to generate a returning boomerang flight. In the demonstrated returning trial, the boomerang is released at $51~\mathrm{rad/s}$ ($8.1~\mathrm{rev/s}$), reaches $2.03~\mathrm{m}$ from the robot base, and returns to touch down $0.31~\mathrm{m}$ from the base. The successful release differs significantly from the measured human throws, showing that a robot need not imitate human throwing motion to achieve a returning flight.  The project page is available at \url{https://robot-boomerang.github.io}.
\end{abstract}

\setlength{\textfloatsep}{8pt}
\section{Introduction}

A returning boomerang converts the kinetic energy of a throw into sustained aerodynamic flight. Released with sufficient forward speed and spin, its rotating blades generate lift, while aerodynamic moments acting on the spinning body cause gyroscopic precession. These effects bend the trajectory into a loop and bring the boomerang back toward the thrower~\cite{hess1975,vassberg2012}. Unlike an ideal ballistic projectile, however, the flight depends not only on translational velocity, but also strongly on the attitude and spin at release.

Beyond recreation, this return capability has been proposed for aerial sensing without onboard propulsion. For example, a boomerang could be launched from a surface platform or balloon on Venus or Titan, collect measurements during flight, and return toward the launcher~\cite{disantis2024}. Such applications require a dedicated mechanism that can produce the release conditions needed for returning flight.

\begin{figure}[t!]
    \centering
    \includegraphics[width=0.95\linewidth]{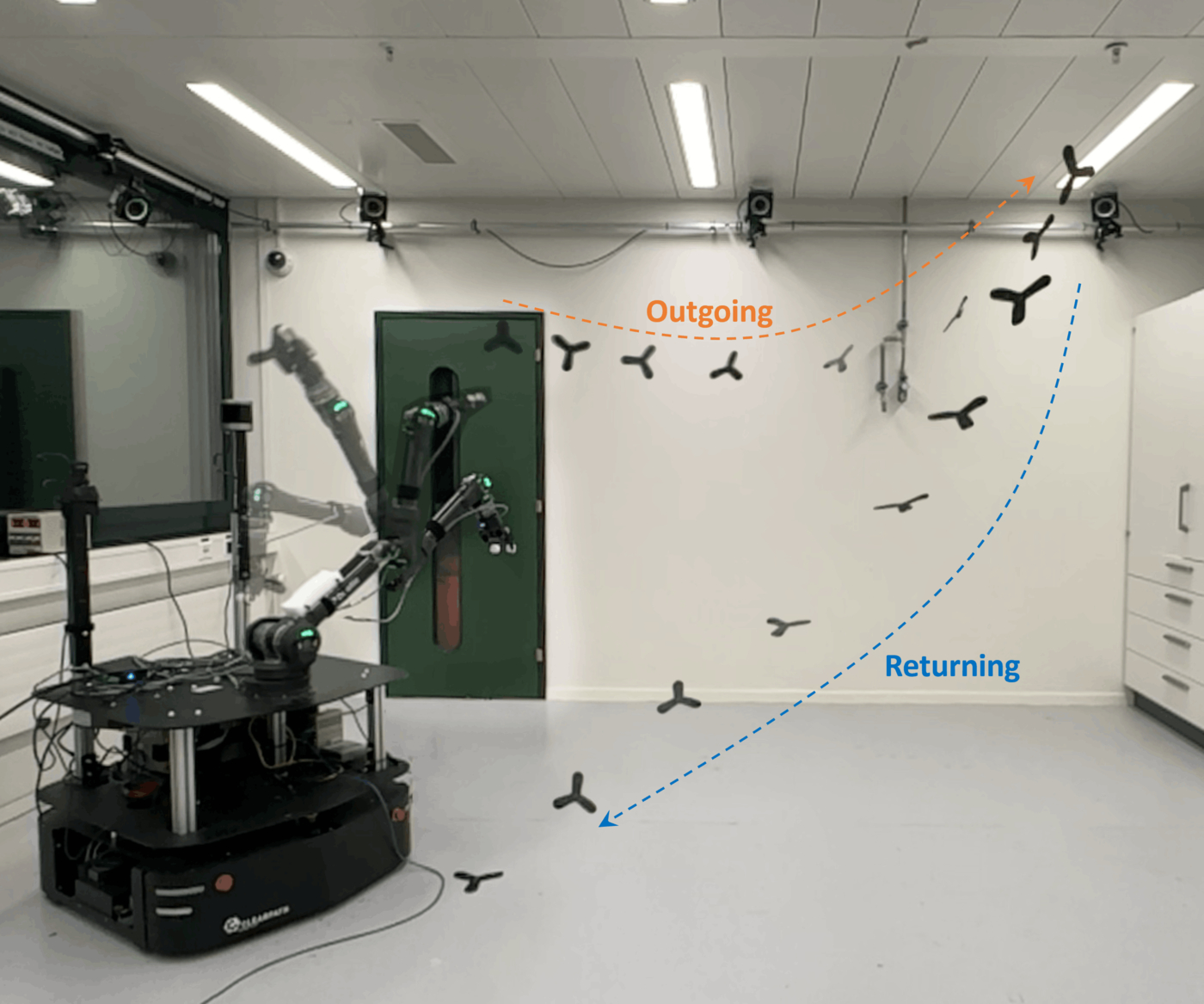}
    \caption{\small Robot throws a boomerang with a returning flight.}
    \label{fig:flight-snapshot}
\end{figure}

In this work, we ask whether a robotic manipulator can generate such a returning flight despite having very different motion capabilities from a human thrower. We show that it can. To our knowledge, this is the first robotic manipulator to produce a returning boomerang flight (Fig.~\ref{fig:flight-snapshot}).

There are two main challenges in designing a returning throw for a robot.

\textbf{Challenge 1: Release and flight dynamics.}
The boomerang does not simply leave the gripper with the motion of the robot fingertips. During the approximately $100~\mathrm{ms}$ gripper-opening interval, the normal force decreases and the boomerang can pivot and slide between the fingers. Friction during this transient changes its motion, so the velocity and spin at detachment can differ substantially from those of the fingertip~\cite{liu2025transient}. After detachment, the problem changes from contact dynamics to aerodynamic flight. The boomerang undergoes six-degree-of-freedom rigid-body motion, while its rapidly rotating blades operate at low Reynolds number and periodically experience reversed flow. Aerodynamic moments also change the attitude through gyroscopic precession, coupling the attitude evolution to the flight path~\cite{hess1975}. Predicting a throw from robot motion therefore requires models for both the transient release and the subsequent flight.

\textbf{Challenge 2: Robot speed limits.}
Returning boomerangs require fast throwing motion and high release spin. Measurements of rapid human overhand throwing report joint angular velocities above $70~\mathrm{rad/s}$ and angular accelerations on the order of $10^3~\mathrm{rad/s^2}$~\cite{roach2013elastic,roach2014upper}. Robotic manipulators operate under different actuator limits. Gear reduction increases output torque at the expense of joint speed, while motor back electromotive force further limits speed during rapid motion. Directly reproducing human throwing motions is therefore generally infeasible. Instead, the release must be generated using the motion that the robot can achieve.

Facing these challenges, we propose a model-based design procedure centered on the release state. A transient release model maps the robot throwing motion to the boomerang state at detachment, while a flight model maps the release state and boomerang design to the subsequent flight. To guide boomerang design, we identify the flight dynamics in stages to predict how the flight changes with the boomerang design. To achieve the high release spin required for return, we systematically screen motion parameters according to their effect on release spin using the release model.

Together, these models guide the design of the throwing motion and boomerang for a 6-DoF manipulator with limited joint speeds. The resulting robot throw operates in a substantially different aerodynamic regime from the measured human throws, showing that a returning throw can be designed for the robot rather than imitating human motion.

\section{Related Work}
\textbf{Robot throwing of complex objects.}
Over the past four decades, robot throwing research has progressively expanded both the variety of objects that can be thrown and the outcomes that can be controlled. Early systems considered nonprehensile throwing of specially shaped objects released from an open palm~\cite{lynch1999dynamic}, as well as in-hand throwing and catching with high-speed multifingered hands~\cite{furukawa2006dynamic}. Prehensile throwing later extended the object set to arbitrary and deformable items, using either large-scale real-world throwing data~\cite{zeng2020tossingbot} or zero-shot robust throwing with model-based motion synthesis~\cite{liu2024tube}. More recently, robot throwing has also considered control of object rotation and landing pose~\cite{liu2025throwflip,dawne2026flipitright}. Across these works, the main modeling challenge lies in the robot--object interaction before detachment, while the subsequent flight is typically described by projectile dynamics under gravity, sometimes with simple air drag.

Throwing aerodynamic objects extends robot throwing beyond ballistic flight because lift can continue to shape the trajectory after release. Aslam et al.~\cite{aslam2025dartbot} throw darts using learned corrections for a given set of dart designs, while Obayashi et al.~\cite{obayashi2023robotic} use hundreds of throws with a fixed robot motion to optimize the shape of paper airplanes. Boomerang throwing couples release and flight more strongly: the release velocity, attitude, and spin determine the subsequent aerodynamic flight, while the high release spin needed for a returning flight is difficult to generate under the joint-speed limits of robotic manipulators. We therefore consider the boomerang design together with the throwing motion that launches it.

\textbf{Boomerang flight dynamics.}
Though primarily used for recreational sport, the boomerang's unusual flight dynamics have attracted a substantial body of aerodynamic research. Hess~\cite{hess1975} combined wind-tunnel force and torque measurements of rotating boomerangs with numerical integration of spin-averaged rigid-body dynamics. Azuma et al.~\cite{azuma2004flight} developed an analytical blade-element model of the aerodynamic coefficients, while Vassberg~\cite{vassberg2012} derived a closed-form approximation of the circular flight radius under steady-flight assumptions. Gudem and colleagues further refined the aerodynamic modeling of reversed flow over the retreating blade~\cite{gudem2019}, with wind-tunnel validation of the predicted lift~\cite{tahmassebpur2021}.

Smart boomerangs have also been proposed as passive returning probes for planetary exploration~\cite{disantis2024}. In hardware tests, hand throws of the prototype reached speeds below $18~\mathrm{m/s}$ and spins of about $2~\mathrm{rev/s}$, but did not produce a returning flight. A purpose-built crossbow launcher could not independently control release timing, speed, and spin, leaving the generation of suitable launch conditions as a key challenge.

Overall, the boomerang literature has focused primarily on how aerodynamics and geometry determine flight, while treating the launch condition as given. Human throwing techniques, e.g.~\cite{broadbent2018}, are difficult to reproduce on robotic manipulators because of their different kinematics and dynamic limits. Robotic boomerang throwing therefore requires linking the flight dynamics to release conditions that the robot can actually generate.

\section{Model-Based Release Design}
\label{sec:method}

The release state links the robot motion to the subsequent boomerang flight. Section~\ref{sec:flight} models how the release state and boomerang design determine the flight. Section~\ref{sec:release} considers how the robot can generate the high release spin required for a returning flight. Section~\ref{sec:method_procedure} combines the two into a design procedure.

\subsection{Spin-averaged Boomerang Flight Dynamics}
\label{sec:flight}

\noindent\textbf{Why a boomerang returns.}
A returning boomerang can be approximated as a spinning disk that generates aerodynamic lift. When released nearly vertical, with initial bank angle $\beta_0 \approx 0$ ($\beta=0$ denotes a vertical disk), much of the lift acts sideways and bends the flight path. The advancing blade also experiences a higher airspeed than the retreating blade, producing an aerodynamic rolling moment. On the rapidly spinning boomerang, this moment causes gyroscopic precession and turns the spin axis together with the flight path. As the disk gradually lays over, part of the lift rotates upward to oppose gravity. For the near-vertical release considered here, a returning flight therefore requires sufficient forward speed and spin.

\begin{figure}[h!]
    \centering
    \includegraphics[width=0.95\linewidth]{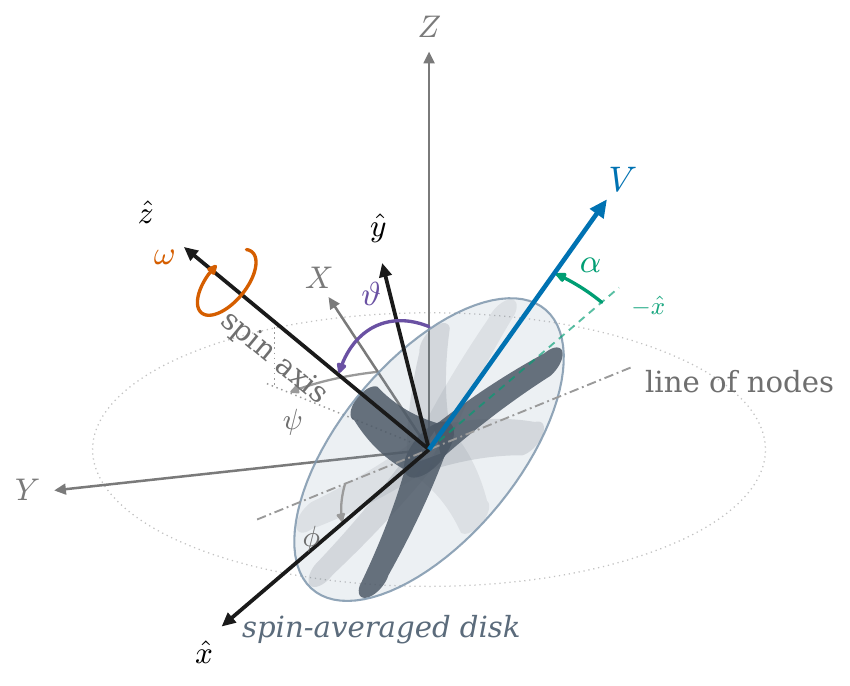}
    \caption{\small Boomerang frames and variable definitions.}
    \label{fig:flight-frames}
\end{figure}

\noindent\textbf{Spin-averaged flight dynamics.}
We use the spin-averaged model of Hess~\cite{hess1975}, which approximates the rapidly rotating boomerang as a spinning disk of mass $m$, effective radius $R$, and spin-axis inertia $I_3$ (Fig.~\ref{fig:flight-frames}). Here, spin averaging suppresses the fast variation of the aerodynamic loads with blade azimuth over each revolution, while retaining the slower evolution of the boomerang velocity, attitude, and spin during flight.

The pose is described by the lab-frame position $(X,Y,Z)$ and the ZXZ Euler angles: precession $\phi$, nutation $\vartheta$, and spin angle $\psi$. The spin rate about the disk axis is denoted by $\omega$. The bank angle is $\beta = 90^\circ-\vartheta$, so $\beta=0$ corresponds to a vertical disk. The translational velocity is described by its magnitude $V$ and the disk angle of attack $\alpha$.

The advance ratio $
U = \frac{V}{\omega R}$ 
compares the translational speed with the characteristic blade-tip speed $\omega R$. At small $U$, the blade-relative airflow is dominated by rotation; as $U$ increases, the difference between the advancing- and retreating-blade airspeeds becomes more pronounced. Together with the disk angle of attack $\alpha$, which determines the incidence of the rotating disk to the incoming flow, this governs the spin-averaged aerodynamic force and torque.

The lift and drag generated by the rotating blades are averaged over a spin cycle and resolved into resultant body-frame forces and torques. We represent these loads using the nondimensional coefficient functions $c_i^F(U,\alpha)$ and $c_i^T(U,\alpha)$. Taking $\omega R$ as the characteristic velocity and $R$ as the characteristic length gives the force scale $\rho(\omega R)^2R^2=\rho R^4\omega^2$ and the corresponding torque scale $\rho R^5\omega^2$,
\begin{equation}
F_i^{\mathrm{aero}}
=
c_i^F(U,\alpha)\rho R^4\omega^2,
\qquad
T_i
=
c_i^T(U,\alpha)\rho R^5\omega^2,
\label{eq:bum-scaling}
\end{equation}
where $i\in\{x,y,z\}$ denotes the body axes and $\rho$ is the air density. The coefficient functions capture the remaining dependence of the aerodynamic loads on the boomerang geometry and operating condition $(U,\alpha)$.

Gravity is projected into the body frame and added to the aerodynamic forces:
\begin{equation}
\begin{bmatrix}
F_x\\
F_y\\
F_z
\end{bmatrix}
=
\begin{bmatrix}
F_x^{\mathrm{aero}}\\
F_y^{\mathrm{aero}}\\
F_z^{\mathrm{aero}}
\end{bmatrix}
-
mg
\begin{bmatrix}
\sin\vartheta\sin\psi\\
\sin\vartheta\cos\psi\\
\cos\vartheta
\end{bmatrix}.
\label{eq:gravity-body}
\end{equation}
Thus, $F_i$ below denotes the total body-frame force, while $T_i$ denotes the aerodynamic torque. The spin-averaged flight dynamics are
\begin{align}
\dot V &= \tfrac{1}{m}\left(-F_x \cos\alpha - F_z \sin\alpha\right) \label{eq:hess-V}\\
\dot \alpha &= \tfrac{1}{mV}\left(F_x \sin\alpha - F_z \cos\alpha\right) + \tfrac{T_x}{I_3 \omega} \label{eq:hess-alpha}\\
\dot \omega &= \tfrac{T_z}{I_3} \label{eq:hess-omega}\\
\dot \vartheta &= -\tfrac{1}{I_3 \omega}\left(T_y \cos\psi + T_x \sin\psi\right) \label{eq:hess-nut}\\
\dot \phi &= \tfrac{1}{I_3 \omega \sin\vartheta}\left(T_x \cos\psi - T_y \sin\psi\right) \label{eq:hess-spin}\\
\dot \psi &= -\tfrac{F_y}{mV\cos\alpha} - \tan\alpha\,\tfrac{T_y}{I_3 \omega} - \cos\vartheta\,\dot\phi. \label{eq:hess-prec}
\end{align}
The position kinematics $(\dot X,\dot Y,\dot Z)$ follow from resolving $V$ through the boomerang attitude.

The flight model separates the nondimensional aerodynamic coefficient functions from the explicit dimensional parameters $m$, $R$, and $I_3$. The tested variants retain the same three-wing layout and flap configuration: the thin variant changes only the blade thickness, while the large variant uniformly scales the in-plane geometry. We therefore make the simplifying assumption that the aerodynamic coefficient functions remain unchanged across the three variants, while updating $m$, $R$, and $I_3$ for each design. This neglects changes in the aerodynamic coefficients caused by the thickness and scale modifications; Section~\ref{subsec:flight-validation} evaluates whether this approximation is sufficient to predict the direction of the resulting flight changes.

\noindent\textbf{Practical model identification.}
Direct measurement of the aerodynamic coefficient functions would require a rotating rig with force--torque sensing in a wind tunnel~\cite{tahmassebpur2021}. We instead identify them from measured flight trajectories.

Jointly fitting all aerodynamic coefficient functions from flight trajectories is difficult because Eqs.~\eqref{eq:hess-V}--\eqref{eq:hess-prec} are nonlinear and coupled. Errors in the estimated coefficients change the simulated velocity and attitude, which in turn change the later aerodynamic conditions $(U,\alpha)$. For the lightweight boomerangs considered here ($4$--$50\,\mathrm{g}$), this makes the fit sensitive to trajectory measurement errors. We therefore first identify a part of the dynamics that can be estimated independently.

The spin dynamics provide this separation. Combining
Eqs.~\eqref{eq:bum-scaling} and~\eqref{eq:hess-omega} gives
\begin{equation}
c_z^T(U,\alpha)
=
\frac{I_3\dot{\omega}}
{\rho R^5\omega^2}.
\label{eq:spin-coeff-direct}
\end{equation}
Thus, along a measured flight trajectory, the spin-axis aerodynamic
coefficient can be estimated directly from the measured flight state,
without integrating the coupled attitude dynamics. We first fit $c_z^T(U,\alpha)$ as a 10-parameter polynomial from the measured spin dynamics. We then retain this fitted function, use the aerodynamic prior for the other five coefficient functions, and fit one multiplicative gain for each of the six functions, giving 16 parameters in total. Section~\ref{subsec:flight-validation} evaluates its flight reconstruction and transfer to other boomerang designs.

\subsection{Finding Robot Throwing Motions for High Release Spin}
\label{sec:release}

\begin{figure}[h!]
    \centering
    \includegraphics[width=0.95\linewidth,
    trim={1.5cm 2.0cm 1.5cm 0.0cm}, 
    clip
    ]
    {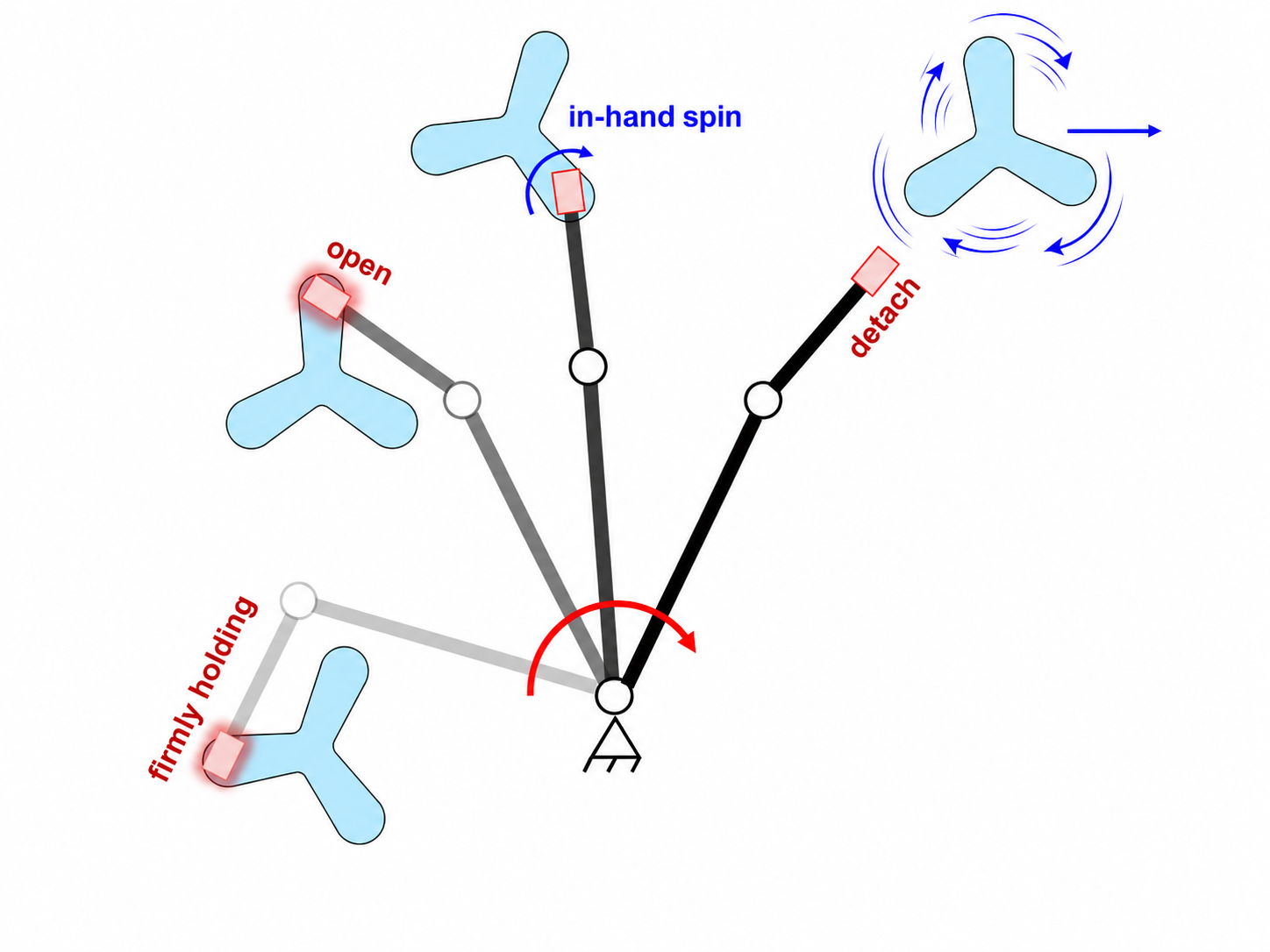}
    \caption{\small Transient release during an overhead throw. The boomerang is initially pinch-grasped. As the gripper opens and the normal force decreases, it pivots and slides in hand, gaining spin before detachment.}
    \label{fig:release-sequence}
\end{figure}

Consider a parallel gripper pinch-grasping the boomerang during the throw. After the opening command, the fingers separate and the contact normal force decreases over approximately $100\,\mathrm{ms}$ before vanishing at detachment. As the contact can no longer maintain sticking, the boomerang pivots and slides between the fingers~\cite{liu2025transient}. This short in-hand motion can substantially increase its spin before release. Figure~\ref{fig:release-sequence} illustrates this process during the overhead throw.

We model such transient release using the Sliding Pivot model of~\cite{liu2025transient}. Given the robot motion, normal-force profile, object and grasp properties, and contact parameters, the model predicts the boomerang state at detachment. Its prediction accuracy is close to that of a more detailed patch-friction model, while simulation is over $20\times$ faster. This makes it ideal for screening many candidate throwing motions.

We use the model to identify which motion parameters provide useful control over the release spin, rather than to optimize the complete throwing motion.

\noindent\textbf{Screening procedure.}
Robot throwing motion admits many possible parameterizations, but only a subset may provide useful leverage over the release spin. Rather than optimizing the complete motion jointly, we consider a set of physically interpretable candidate parameters $u=(u_1,\ldots,u_m)$ and use the Sliding Pivot model to screen how effectively each can modulate the predicted release spin. For a candidate-parameter setting $u$ and uncertain operating condition $\xi\in\mathcal X$, covering variation in the robot release state, normal-force profile, object and grasp properties, and contact parameters, the model predicts the release spin $\omega_0(u,\xi)$.

To screen candidate $u_i$, we evaluate $K$ uniformly spaced levels over its prescribed range while fixing all other candidate parameters at a common operating point, and reuse the same sampled operating conditions for every candidate sweep. Let $u^{(i,j)}$ denote the complete candidate-parameter vector at level $j$, with only its $i$th component varied. For each operating condition $\xi$ and $j=0,\ldots,K-2$, the normalized local change in release spin between adjacent levels is

\begin{equation}
g_{i,j}(\xi)=(K-1)\left[\omega_0\!\left(u^{(i,j+1)},\xi\right)-\omega_0\!\left(u^{(i,j)},\xi\right)\right].
\label{eq:local-gain}
\end{equation}

The factor $K-1$ normalizes each adjacent change by the fraction of the full parameter range traversed between two neighboring levels.

We summarize the resulting local effects using the \emph{spin authority} and \emph{consistency},
\begin{align}
A_i &= \left|\mathbb{E}_{\xi,j}[g_{i,j}]\right|,\qquad c_i=\frac{\left|\mathbb{E}_{\xi,j}[g_{i,j}]\right|}{\sqrt{\mathbb{E}_{\xi,j}[g_{i,j}^2]}}\in[0,1].
\label{eq:screening}
\end{align}
Here, $\mathbb{E}_{\xi,j}$ denotes the empirical mean over the sampled operating conditions and adjacent sweep levels. Because the sweep is uniform, $A_i$ equals the magnitude of the mean endpoint change in release spin across the operating conditions, and therefore measures how much candidate $u_i$ can modulate the release spin over its tested range. The consistency $c_i$ compares the signed mean local effect with its RMS magnitude: it is high when the effect maintains a similar direction and magnitude across the sweep and operating conditions, and decreases when the effect varies strongly or changes sign. Taking the absolute value after averaging preserves this sign cancellation. Section~\ref{sec:exp_screening} specifies the candidate parameters, operating-condition ranges, and sampling used in the experiments.

\subsection{Design Procedure}
\label{sec:method_procedure}

The flight and release models are combined through the release state, with the design ordered by iteration cost: a throwing motion can be changed and tested within seconds, whereas manufacturing a new boomerang takes hours. We therefore begin with an initial boomerang and a reference throwing motion, and evaluate the resulting flight.

We first adjust the release spin using the motion parameters identified in Sec.~\ref{sec:release}. If the boomerang turns too little, the spin is increased within the tested motion family; if it turns too strongly, the spin can be reduced. If the boomerang still turns insufficiently at the highest achievable spin within this motion family, further motion tuning is no longer sufficient and the boomerang design is changed.

We then use the flight model of Sec.~\ref{sec:flight} differentially: under comparable release conditions, it predicts how a change in boomerang design will change the subsequent trajectory. This allows the design change to be selected from the predicted direction of the flight response, rather than requiring the model to predict an absolute returning release. The revised boomerang is then tested using the high-spin throwing motion, and the procedure is repeated as needed.

\section{Experiments}

We evaluate the framework in three stages. Section~\ref{subsec:flight-validation} identifies the flight model from hand-thrown trajectories and assesses its reconstruction and transfer across boomerang designs. Section~\ref{sec:exp_screening} screens candidate motion parameters in simulation and validates their effects on robot release spin. Section~\ref{sec:exp-return} then combines the two models to obtain a returning flight.

\begin{figure}[h!]
    \centering
    \includegraphics[
        width=0.9\columnwidth,
        trim={1.5cm 1.5cm 1.5cm 1.0cm},
        clip
    ]{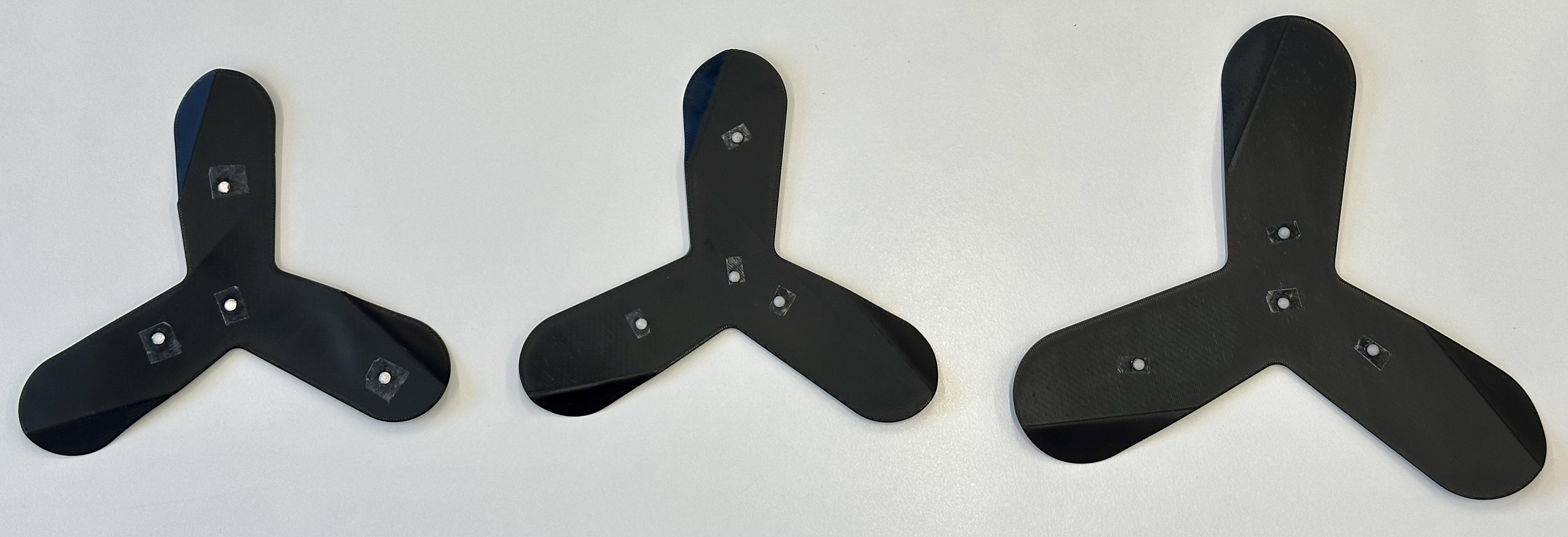}

     \caption{\small Three tested boomerangs. Left to right: nominal ($0.6\,\mathrm{mm}$ thick), thin ($0.4\,\mathrm{mm}$), and large ($0.6\,\mathrm{mm}$).}
    \label{fig:variants}
\end{figure}

\noindent\textbf{Shared setup.}
Boomerang motion is recorded with an OptiTrack motion-capture system at 300~Hz. A marker cluster provides the release state and full 6-DoF trajectory. We test three 3D-printed variants (Fig.~\ref{fig:variants}, Table~\ref{tab:variants}): 
the nominal design, a thin variant with unchanged in-plane geometry, and a large variant with the in-plane geometry scaled by $1.25\times$. The nominal variant is used for model identification, and the thin and large variants for cross-design transfer.

\begin{table}[h!]
\centering
\caption{\small Boomerang variants used in the experiments.}
\label{tab:variants}
\footnotesize
\setlength{\tabcolsep}{4pt}
\begin{tabular}{@{}lccccc@{}}
\hline
variant &  thickness (mm) & radius (mm) & $m$ (g) & $I_3$ ($\mathrm{g\,cm^2}$) \\
\hline
nominal & 0.6 & 85.00  & 6.000 & 1.4617 \\
thin    & 0.4 & 85.00  & 4.000 & 0.9745 \\
large   &  0.6 & 106.25 & 9.375 & 3.5686 \\
\hline
\end{tabular}
\end{table}

\subsection{Flight-model identification and cross-design assessment}
\label{subsec:flight-validation}

\noindent\textbf{Hand-thrown data.}
We identify the flight model from $N=9$ hand-thrown flights of the nominal boomerang. Each simulation is initialized from the measured release state of the corresponding trial. Across the nine flights, $V_0\in[5.1,6.1]\,\mathrm{m/s}$ and $\omega_0\in[37.4,51.9]\,\mathrm{rad/s}$.

\noindent\textbf{Flight reconstruction.}
\label{sec:exp-recon}
To evaluate the staged identification of Sec.~\ref{sec:flight}, we compare four model classes on the same nine flights. The \emph{wind-tunnel prior} uses the published wind-tunnel coefficient functions of~\cite{hess1975} without fitting (0 parameters), testing direct transfer. The \emph{scaled wind-tunnel prior} fits one multiplicative gain for each of the six coefficient functions in Eq.~(1) (6 parameters), testing whether correcting their magnitudes is sufficient. The \emph{direct polynomial fit} fits all six coefficient functions independently as polynomials in $(U,\alpha)$ (60 parameters), testing identification without an aerodynamic prior. The \emph{staged fit (ours)} first identifies $c_z^T(U,\alpha)$ from the measured spin dynamics and then fits six scalar gains while retaining the wind-tunnel structure for the other five coefficient functions (16 parameters).

The staged fit is structurally better posed than fitting all six coefficient functions directly: $\dot{\omega}=T_z/I_3$ constrains the spin-axis torque from the measured spin dynamics without integrating the coupled attitude dynamics, while the remaining five aerodynamic channels retain the prior structure. For the fitted models, the same nine flights are used for calibration and reconstruction, so this is an in-sample reconstruction assessment rather than a held-out test. Table~\ref{tab:ablation} reports whole-trajectory dynamic time warping (DTW), landing-position error, and apex-height error as MAE $\pm$ STD.

\begin{table}[t!]
\centering
\caption{\small Flight-reconstruction error by model class on the nominal boomerang (MAE $\pm$ STD over $N{=}9$ trials). Boldface indicates the lowest MAE for each metric.}
\label{tab:ablation}
\footnotesize
\setlength{\tabcolsep}{3.5pt}
\begin{tabular}{@{}lcccc@{}}
\hline
class & \#par. & DTW (m) & landing (m) & $Z_{\max}$ (m) \\
\hline
wind-tunnel prior & 0 & 0.63 $\pm$ 0.18 & 2.09 $\pm$ 0.50 & 0.46 $\pm$ 0.11 \\
\shortstack[l]{scaled wind-tunnel\\prior} & 6 & 0.31 $\pm$ 0.13 & \textbf{0.37 $\pm$ 0.23} & 0.27 $\pm$ 0.09 \\
\shortstack[l]{direct polynomial\\fit} & 60 & 29 $\pm$ 60 & 80 $\pm$ 163 & 12 $\pm$ 28 \\
staged fit (ours) & 16 & \textbf{0.29 $\pm$ 0.19} & 0.55 $\pm$ 0.38 & \textbf{0.08 $\pm$ 0.06} \\
\hline
\end{tabular}
\end{table}

\begin{figure}[h!]
    \centering
    \includegraphics[width=0.9\linewidth]{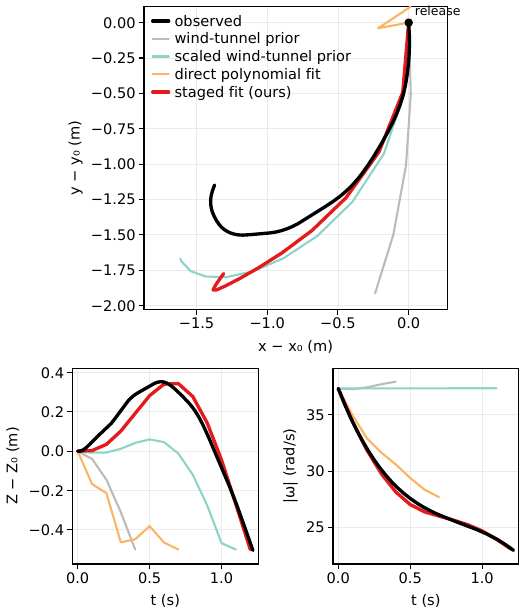}
    \caption{\small Reconstruction of a representative nominal flight by model class: top view (top), height above release (bottom left), and spin rate (bottom right).}

    \label{fig:recon-overlay}
\end{figure}

Table~\ref{tab:ablation} and Fig.~\ref{fig:recon-overlay} show the main failure modes. The \emph{wind-tunnel prior} transfers poorly, while the \emph{direct polynomial fit} becomes unstable during forward integration. The \emph{scaled wind-tunnel prior} gives the lowest landing error but under-predicts the apex height on all nine flights, with a mean bias of $-0.27\,\mathrm{m}$. The \emph{staged fit (ours)} gives the lowest DTW and apex-height error, with only $+0.01\,\mathrm{m}$ mean apex bias, and is therefore used below.

\noindent\textbf{Design sensitivity at matched release.}
\label{sec:exp-delta}
Because Sec.~\ref{sec:method_procedure} uses the flight model differentially, we next test whether it predicts the direction of design-induced flight changes without refitting. The staged fit is identified only on the nominal boomerang and transferred to the thin and large variants with the coefficient functions unchanged, while $m$, $R$, and $I_3$ are updated according to Table~\ref{tab:variants}.

\begin{figure}[h!]
    \centering
    \includegraphics[width=0.8\linewidth]{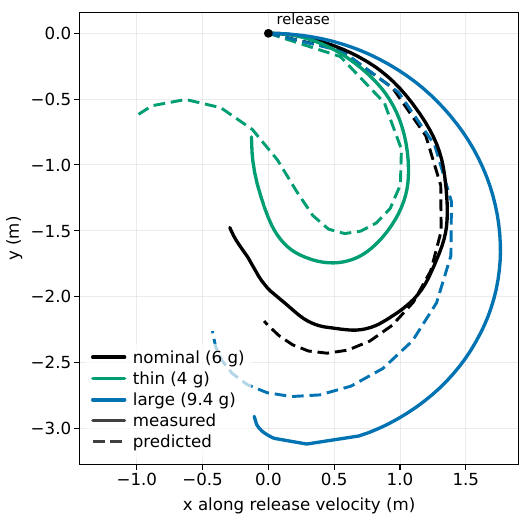}
    \caption{\small Measured and predicted cross-design trajectories at closely matched release states in a release-aligned frame. Relative to the nominal design, the thin variant turns more tightly and the large variant more widely; the transferred staged fit predicts both changes without refitting.}
    \label{fig:delta-overlay}
\end{figure}

Because hand throwing cannot reproduce identical releases, we select one closely matched throw from each variant; across the three releases, speed differs by at most $3.3\%$, spin by $2.0\%$, and attitude by $5.8^\circ$. We quantify horizontal turning by $R_{\mathrm{curv}}$, the arc-length-averaged radius of curvature over the first $2\,\mathrm{m}$ of free flight. Figure~\ref{fig:delta-overlay} compares the measured and predicted trajectories.

The measured $R_{\mathrm{curv}}$ changes from $1.22\,\mathrm{m}$ for the nominal design to $0.88\,\mathrm{m}$ for the thin variant and $1.58\,\mathrm{m}$ for the large variant; the model predicts $1.25$, $0.76$, and $1.36\,\mathrm{m}$, respectively. Thus, although it does not predict the exact change in curvature, it correctly predicts the direction of both design changes without refitting. This is the differential capability used in Sec.~\ref{sec:exp-return} to guide the boomerang modification.

\subsection{Motion Screening and Release-Spin Validation}
\label{sec:exp_screening}

\noindent\textbf{Hardware setup.}
The robot experiments use a DynaArm, a torque-controlled 6-DoF manipulator with $1.1\,\mathrm{m}$ reach, equipped with a Robotiq 2F-85 parallel gripper and controlled at $500\,\mathrm{Hz}$. The arm has a maximum end-effector speed of $\sim$$10\,\mathrm{m/s}$, with peak elbow and wrist speeds of $7$ and $14\,\mathrm{rad/s}$, respectively. These actuator limits make it difficult to generate the high release spin required for returning boomerang flight directly from robot joint motion, motivating the use of transient in-hand rotation and the search for effective motion parameters.

\begin{figure}[h!]
    \centering
    \includegraphics[
        width=0.43\columnwidth,
        trim={0.0cm 5.0cm 0.0cm 1.0cm},  
        clip
    ]{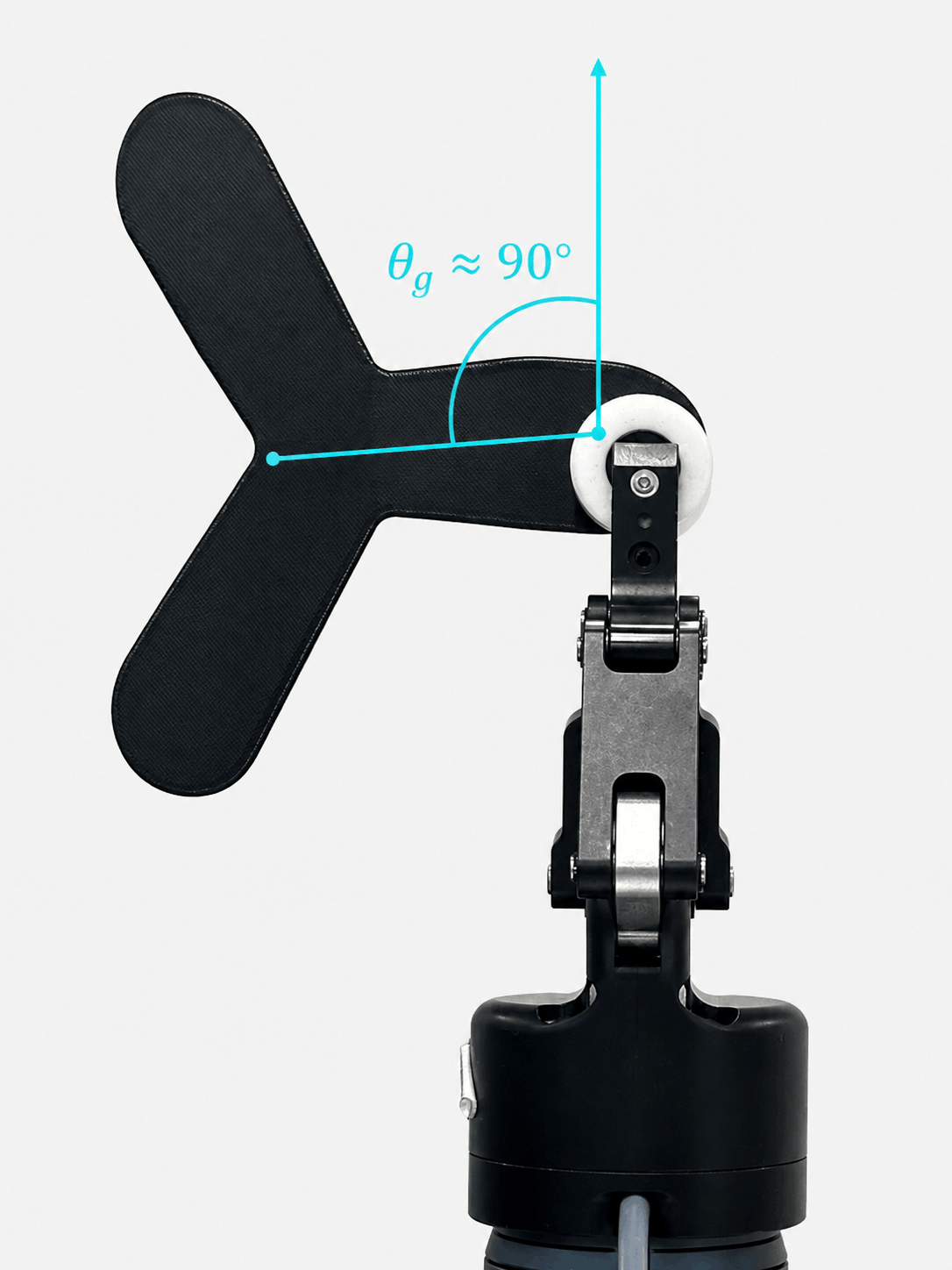}
    \hspace{0.02\columnwidth}
    \includegraphics[
        width=0.43\columnwidth,
        trim={0.0cm 5.0cm 0.0cm 1.5cm},  
        clip
    ]{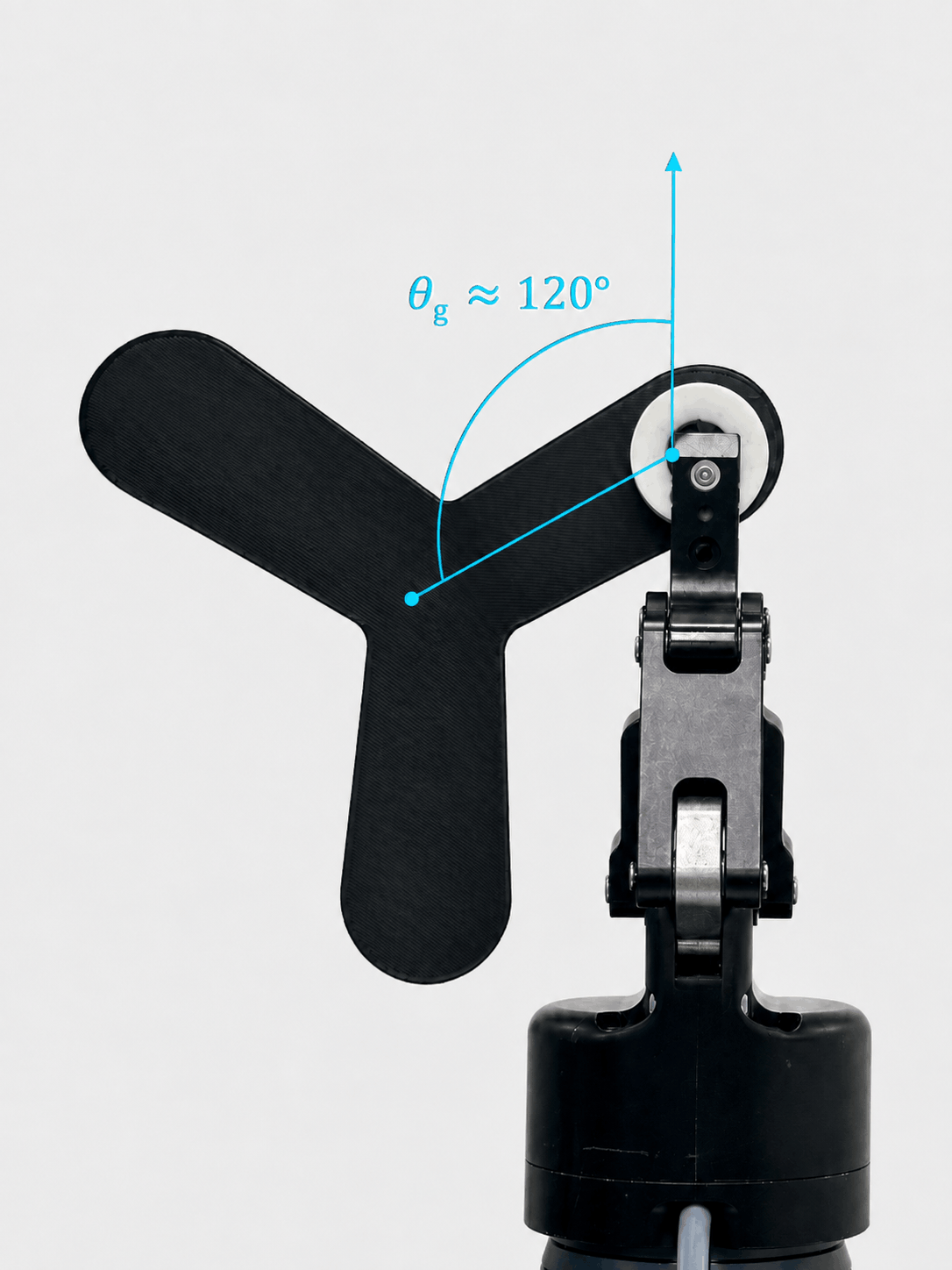}
    \caption{\small Illustration of the grasp angle $\theta_g$.}
    \label{fig:grasp_angle}
\end{figure}

\noindent\textbf{Candidate motion parameters.}
We consider four candidate parameters for controlling the release spin:
1) grasp angle $\theta_g \in [0,120]^\circ$, illustrated in Fig.~\ref{fig:grasp_angle};
2) terminal elbow deceleration $a_e \in [0,30]\,\mathrm{rad/s^2}$;
3) terminal wrist deceleration $a_w \in [0,60]\,\mathrm{rad/s^2}$; and
4) gripper opening delay $t_d \in [0.03,0.08]\,\mathrm{s}$.
The deceleration spans are determined by the actuator limits, while the grasp-angle and delay spans cover the ranges usable on the experimental setup. We first screen these parameters in simulation using the release model of Sec.~\ref{sec:release}, then test the predicted release-spin responses on the robot.

\noindent\textbf{Simulation screening.}
Each parameter is swept over 10 uniformly spaced levels while the other three are fixed at a common operating point: $\theta_g=90^\circ$, $a_e=15\,\mathrm{rad/s^2}$, $a_w=30\,\mathrm{rad/s^2}$, and $t_d=0.063\,\mathrm{s}$.

The 12 uncertain operating variables in Table~\ref{tab:nuisance} define $\xi\in\mathcal X\subset\mathbb{R}^{12}$; the same 512 Sobol samples over $\mathcal X$ are reused for every candidate-parameter sweep.
\begin{table}[t]
\caption{\small 12 uncertain operating variables and their prescribed ranges, covering the operating envelope of the setup.}
\label{tab:nuisance}
\centering
\begin{tabular}{llc}
\toprule
group & variable & range \\
\midrule
release state & throw-plane pitch offset [rad] & $[-0.2, 0.2]$ \\
              & elbow release velocity [rad/s] & $[-7, -6]$ \\
              & wrist release velocity [rad/s] & $[-14, -13]$ \\
              & hardware delay [s] & $[0.01, 0.02]$ \\
gripper force & peak force $F_{\max}$ [N] & $[60, 90]$ \\
              & switching force $F_s$ [N] & $[5, 15]$ \\
              & phase-1 duration $T_1$ [s] & $[0.02, 0.03]$ \\
              & phase-2 duration $T_2$ [s] & $[0.02, 0.035]$ \\
grasp, object & grasp offset [m] & $[0.06, 0.10]$ \\
              & object mass [g] & $[4, 30]$ \\
contact       & friction coefficient $\mu$ & $[0.3, 0.8]$ \\
              & contact radius [mm] & $[1, 5]$ \\
\bottomrule
\end{tabular}
\end{table}

\begin{figure}[t!]
    \centering
    \includegraphics[width=0.88\columnwidth]{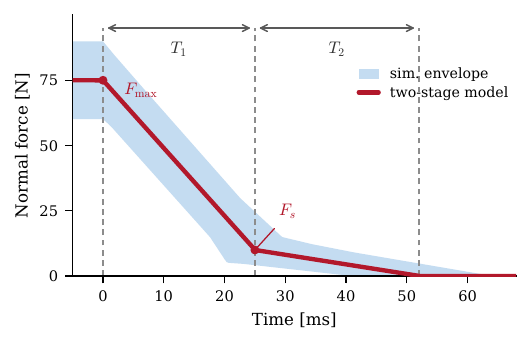}
    \caption{\small Two-stage normal-force profile used in the release simulation. The nominal profile is parameterized by the peak force $F_{\max}$, switching force $F_s$, and phase durations $T_1,T_2$; the shaded region shows the range sampled during screening.}
    \label{fig:normal-force}
\end{figure}

\begin{figure}[t]
\centering
\includegraphics[width=\columnwidth]{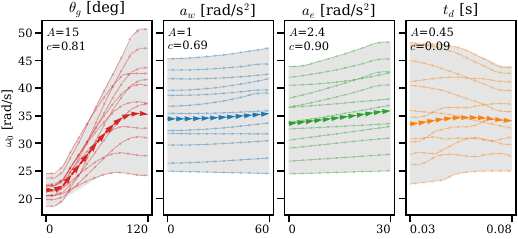}
\caption{\small Simulated release-spin response for the four candidate motion parameters. Each curve shows $\omega_0$ as one parameter is swept under a fixed operating condition; the bold curve is the ensemble mean and the shaded band shows the range. The overall change of the mean reflects spin authority $A$, while consistent local slopes across parameter levels and operating conditions correspond to high consistency $c$. Twelve conditions are shown for legibility.}
\label{fig:screening_strip}
\end{figure}

Table~\ref{tab:screening} reports the spin authority and consistency of each candidate, while Fig.~\ref{fig:screening_strip} shows the simulated release-spin responses. As defined in Sec.~\ref{sec:release}, spin authority measures the magnitude of the mean release-spin change over the tested parameter range, whereas consistency is a normalized measure that decreases when the local effect varies strongly or changes sign across the sweep and uncertain operating conditions.

\begin{table}[t]
\caption{\small Simulation screening over the 512 fixed uncertain operating conditions. $A$: spin authority over the parameter span. $c$: consistency of the local steering gain across parameter levels and operating conditions. $\omega_{\max}$: largest simulated release spin during the sweep.}
\label{tab:screening}
\centering
\begin{tabular}{lcccc}
\toprule
parameter & span & $A$ [rad/s] & $c$ & $\omega_{\max}$ [rad/s] \\
\midrule
$\theta_g$ [deg]      & $[0,120]$     & \textbf{15.0} & 0.81          & \textbf{53.4} \\
$a_e$ [rad/s$^2$]     & $[0,30]$      & 2.43          & \textbf{0.90} & 49.1 \\
$a_w$ [rad/s$^2$]     & $[0,60]$      & 1.00          & 0.69          & 47.3 \\
$t_d$ [s]             & $[0.03,0.08]$ & 0.45          & 0.09          & 48.3 \\
\bottomrule
\end{tabular}
\end{table}

Grasp angle has by far the largest spin authority, $A=15.0\,\mathrm{rad/s}$, while maintaining high consistency ($c=0.81$), and reaches the highest simulated release spin of $53.4\,\mathrm{rad/s}$ ($8.5\,\mathrm{rev/s}$). It is therefore the dominant parameter for producing large changes in release spin. Elbow deceleration has substantially lower spin authority ($A=2.43\,\mathrm{rad/s}$) but the highest consistency ($c=0.90$), while wrist deceleration has still lower spin authority ($A=1.00\,\mathrm{rad/s}$) and consistency ($c=0.69$). These two parameters provide smaller adjustments of release spin.

Gripper delay has both the lowest spin authority ($A=0.45\,\mathrm{rad/s}$) and the lowest consistency ($c=0.09$). Its effect changes strongly across the sweep and operating conditions, making it a poor parameter for predictable release-spin tuning.

\begin{figure}[t]
\centering
\includegraphics[width=\columnwidth]{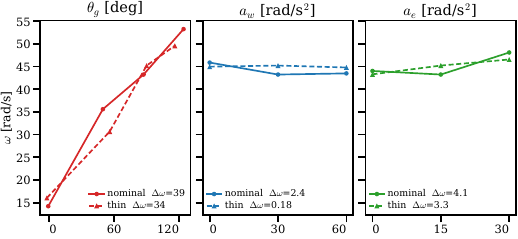}
\caption{\small Measured release-spin response on the robot for the three motion parameters retained after simulation screening. Each curve corresponds to one boomerang variant while one motion parameter is varied and the remaining parameters are held at the operating point.}
\label{fig:knob_strip_hw}
\end{figure}

\noindent\textbf{Robot validation of release spin.}
The simulation identifies gripper delay $t_d$ as the least consistent candidate, so we test the remaining three parameters on the robot. We conduct 16 throws using the nominal and thin boomerangs, varying $\theta_g$, $a_e$, or $a_w$ one at a time around the same operating point. These experiments evaluate the release spin at detachment; the subsequent flight is not used to assess the parameter screening. Figure~\ref{fig:knob_strip_hw} shows the measured release-spin responses corresponding to the simulated sweeps in Fig.~\ref{fig:screening_strip}.

The simulated spin authority ranking,
$\theta_g \gg a_e > a_w$,
is reproduced for both boomerang variants. The measured full-span changes are $39.1$ and $33.6\,\mathrm{rad/s}$ for $\theta_g$, $4.1$ and $3.3\,\mathrm{rad/s}$ for $a_e$, and $2.4$ and $0.2\,\mathrm{rad/s}$ for $a_w$, for the nominal and thin variants, respectively. These measured spans need not equal the simulated spin authority $A$, which averages the endpoint response over the 512 uncertain operating conditions; here, the hardware comparison tests the predicted ranking and response trends.

The largest measured release spin is $53.3\,\mathrm{rad/s}$, obtained at the largest tested grasp angle, close to the simulated maximum of $53.4\,\mathrm{rad/s}$. For both variants, the measured release spin also increases at every tested grasp-angle level, consistent with the strong positive grasp-angle effect predicted in Fig.~\ref{fig:screening_strip}.

Thus, the robot experiments reproduce the main screening result: grasp angle has the strongest effect on release spin, while elbow and wrist decelerations provide smaller adjustments. We use these parameters in the returning-flight experiment of Sec.~\ref{sec:exp-return}.

\subsection{Designing a Returning Throw}
\label{sec:exp-return}
Having validated flight-model transfer in Sec.~\ref{subsec:flight-validation} and release-spin screening in Sec.~\ref{sec:exp_screening}, we apply the design procedure of Sec.~\ref{sec:method_procedure}: first tune the release spin, then modify the boomerang if motion tuning is insufficient. For the throwing motion, we first set the elbow and wrist release velocities to their maximum feasible values ($7$ and $14\,\mathrm{rad/s}$, respectively), limited by motor back-EMF. The release configuration $q$ is then chosen such that the end-effector velocity $J_v(q)\dot q$ is approximately horizontal. Given the target release state $(q,\dot q)$, Ruckig~\cite{berscheid2021jerk} generates a time-optimal trajectory subject to the robot's joint velocity, acceleration, and jerk limits. The obtained throwing trajectory is executed by the joint trajectory controller running at 500 Hz.

\begin{figure}[t]
\centering
\includegraphics[width=\columnwidth]{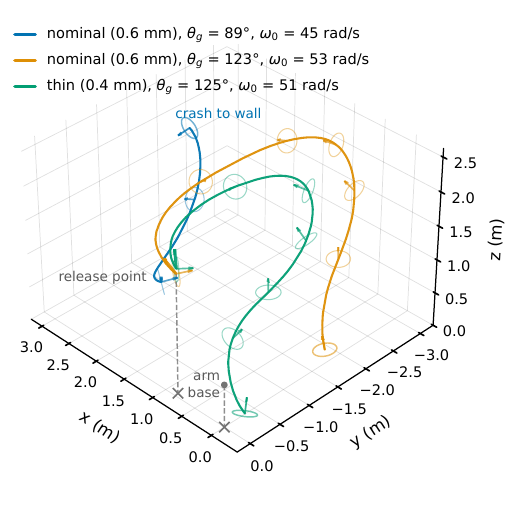}
\caption{\small Measured robot flights illustrating the design procedure. Increasing the grasp angle raises the nominal variant's release spin and produces a clear turn, but the boomerang does not return. The thin variant returns at a similar release spin. For each throw, the release configuration is chosen so that the end-effector velocity is approximately horizontal.}
\label{fig:flights3d}
\end{figure}

Figure~\ref{fig:flights3d} shows three measured robot flights illustrating the progression from increasing release spin to changing the boomerang design. The reported grasp angles are measured with OptiTrack rather than taken from the nominal placement values; small deviations from $90^\circ$ and $120^\circ$ arise from manual placement of the boomerang in the gripper. We define the flight range as the maximum horizontal distance from the robot base and the return distance as the horizontal touchdown distance from the base.

\par\smallskip
\noindent\textbf{\emph{1) Initial throw.}}
We begin with the nominal boomerang at $\theta_g \approx 90^\circ$. The boomerang is released at $\omega_0 = 45\,\mathrm{rad/s}$, follows a nearly straight path, climbs, and strikes the wall before returning.

\par\smallskip
\noindent\textbf{\emph{2) Increase the release spin.}}
Following the screening results of Sec.~\ref{sec:exp_screening}, we increase the grasp angle to near the upper end of the simulated interval $[0,120]^\circ$. With the nominal variant at a measured grasp angle of $\theta_g \approx 123^\circ$, the release spin increases to $\omega_0 = 53\,\mathrm{rad/s}$. The boomerang now turns substantially and lays over during flight, but the turn remains too wide to return to the robot. It reaches a range of $3.2\,\mathrm{m}$ and lands $1.91\,\mathrm{m}$ from the robot base.

\par\smallskip
\noindent\textbf{\emph{3) Change the boomerang design.}}

The nominal variant does not return at the highest release spin achieved within the tested motion family, so we next change its design. From the differential result of Sec.~\ref{subsec:flight-validation}, reducing the blade thickness is predicted to tighten the turn at a comparable release state. We replace the nominal boomerang with the thin variant of Table~\ref{tab:variants} and use the high-spin throwing motion.

At a measured grasp angle of $\theta_g \approx 125^\circ$, the thin variant is released at $\omega_0 = 51\,\mathrm{rad/s}$ ($8.1\,\mathrm{rev/s}$) and completes a returning flight. It reaches a range of $2.03\,\mathrm{m}$, curves back toward the robot, and touches down $0.31\,\mathrm{m}$ from the base after $1.9\,\mathrm{s}$ of free flight ($2.04\,\mathrm{s}$ from the gripper-opening command). 

To evaluate repeatability, we conduct four trials at the final settings. All four trials produce returning flights. Across these trials, the return distance is $0.41 \pm 0.11\,\mathrm{m}$,
the flight range is $2.04 \pm 0.03\,\mathrm{m}$, 
and the flight time is $1.85 \pm 0.04\,\mathrm{s}$. The illustrated trial therefore lies within the variation observed across repeated throws.

\noindent\textbf{Discussion.}
The fastest-spin human and robot throws in our experiments reach similar spin rates ($51.9$ and $53.3\,\mathrm{rad/s}$), but their release speeds differ ($5.95$ and $9.74\,\mathrm{m/s}$), giving advance ratios of $U_0=1.35$ and $2.15$, respectively. Therefore, the human and robot throws operate under very different aerodynamic conditions.

In the returning throw, the boomerang is released at $51\,\mathrm{rad/s}$, while the end-effector angular velocity reaches only $20\,\mathrm{rad/s}$ during release. Thus, the boomerang spins more than $2.5\times$ faster than the end-effector at detachment, showing the substantial contribution of transient in-hand motion to the release spin.

\section{Conclusion}
\label{sec:conclusion}
To our knowledge, we are the first to show a robotic manipulator generating a returning boomerang flight. More importantly, the result shows that dynamic aerodynamic throwing does not require robots to reproduce human throwing motions: by modeling the release state and designing the throwing motion and flying object around it, both can instead be adapted to the robot's own dynamic capabilities. We believe this provides a step towards autonomous throwing--flying systems, where robots launch aerodynamic objects to exploit lift for extended-range transport, sensing, and return.

The current experiments focus on small indoor boomerangs and use aerodynamic coefficients calibrated from measured flights. A natural next step is robotic throwing of larger outdoor boomerangs, where the boomerang geometry can become an explicit design variable. Aerodynamic models based on blade-element theory or CFD could be used to predict the coefficient tables from geometry and guide the design of the airfoil and blade shape together with the release condition.

\bibliographystyle{IEEEtran}
\bibliography{references.bib}

\end{document}